\documentclass[acmtog,nonacm]{acmart}
\makeatletter
\let\@authorsaddresses\@empty
\makeatother
\acmSubmissionID{183}

\usepackage{booktabs} 

\usepackage[ruled]{algorithm2e} 

\SetAlFnt{\small}
\SetAlCapFnt{\small}
\SetAlCapNameFnt{\small}
\SetAlCapHSkip{0pt}

\acmJournal{TOG}

\begin{document}
\title{\textbf{MARS}: \textbf{M}esh \textbf{A}uto\textbf{R}egressive Model for 3D \textbf{S}hape Detailization}

\author{Jingnan Gao}
\affiliation{%
 \institution{Shanghai Jiao Tong University}
 \country{China}
 }
\email{gjn0310@sjtu.edu.cn}
\authornote{The contribution is made during an internship at Tencent XR Vision Labs.} 
\author{Weizhe Liu}
\affiliation{%
 \institution{Tencent XR Vision Labs}
  \country{China}
 }
\email{weizheliu1991@163.com}
 \authornote{Project lead.} 
\author{Weixuan Sun}
\affiliation{%
 \institution{Tencent XR Vision Labs}
  \country{China}
 }
\email{weixuansun7@outlook.com}
\author{Senbo Wang}
\affiliation{%
 \institution{Tencent XR Vision Labs}
  \country{China}
 }
\email{wsb_pro@live.com}
\author{Xibin Song}
\affiliation{%
 \institution{Tencent XR Vision Labs}
  \country{China}
 }
\email{song.sducg@gmail.com}
\author{Taizhang Shang}
\affiliation{%
 \institution{Tencent XR Vision Labs}
  \country{China}
 }
\email{taizhangshang@qq.com}
\author{Shenzhou Chen}
\affiliation{%
 \institution{Tencent XR Vision Labs}
  \country{China}
 }
\email{chenshenzhou@zju.edu.cn}
\author{Hongdong Li}
\affiliation{%
 \institution{Australian National University}
  \country{Australia}
 }
\email{HONGDONG.LI@anu.edu.au}
\author{Xiaokang Yang}
\affiliation{%
 \institution{Shanghai Jiao Tong University}
  \country{China}
 }
 \email{xkyang@sjtu.edu.cn}
\author{Yichao Yan}
\affiliation{%
 \institution{Shanghai Jiao Tong University}
  \country{China}
 }
\email{yanyichao@sjtu.edu.cn}
\authornote{Corresponding author.} 
 \author{Pan Ji}
\affiliation{%
 \institution{Tencent XR Vision Labs}
  \country{China}
 }
 \email{peterji1990@gmail.com}



\begin{abstract}
State-of-the-art methods for mesh detailization predominantly utilize Generative Adversarial Networks (GANs) to generate detailed meshes from coarse ones. These methods typically learn a specific style code for each category or similar categories without enforcing geometry supervision across different Levels of Detail (LODs). Consequently, such methods often fail to generalize across a broader range of categories and cannot ensure shape consistency throughout the detailization process. In this paper, we introduce MARS, a novel approach for 3D shape detailization. Our method capitalizes on a novel multi-LOD, multi-category mesh representation to learn shape-consistent mesh representations in latent space across different LODs. We further propose a mesh autoregressive model capable of generating such latent representations through next-LOD token prediction. This approach significantly enhances the realism of the generated shapes. Extensive experiments conducted on the challenging 3D Shape Detailization benchmark demonstrate that our proposed MARS model achieves state-of-the-art performance, surpassing existing methods in both qualitative and quantitative assessments. Notably, the model's capability to generate fine-grained details while preserving the overall shape integrity is particularly commendable.
\end{abstract}

%
%

\ccsdesc[500]{Computing methodologies ~ Shape modeling.}

%
%

\keywords{Generative model, geometry detailization, autoregressive model, machine learning.}

\begin{teaserfigure}
    \includegraphics[width=\textwidth]{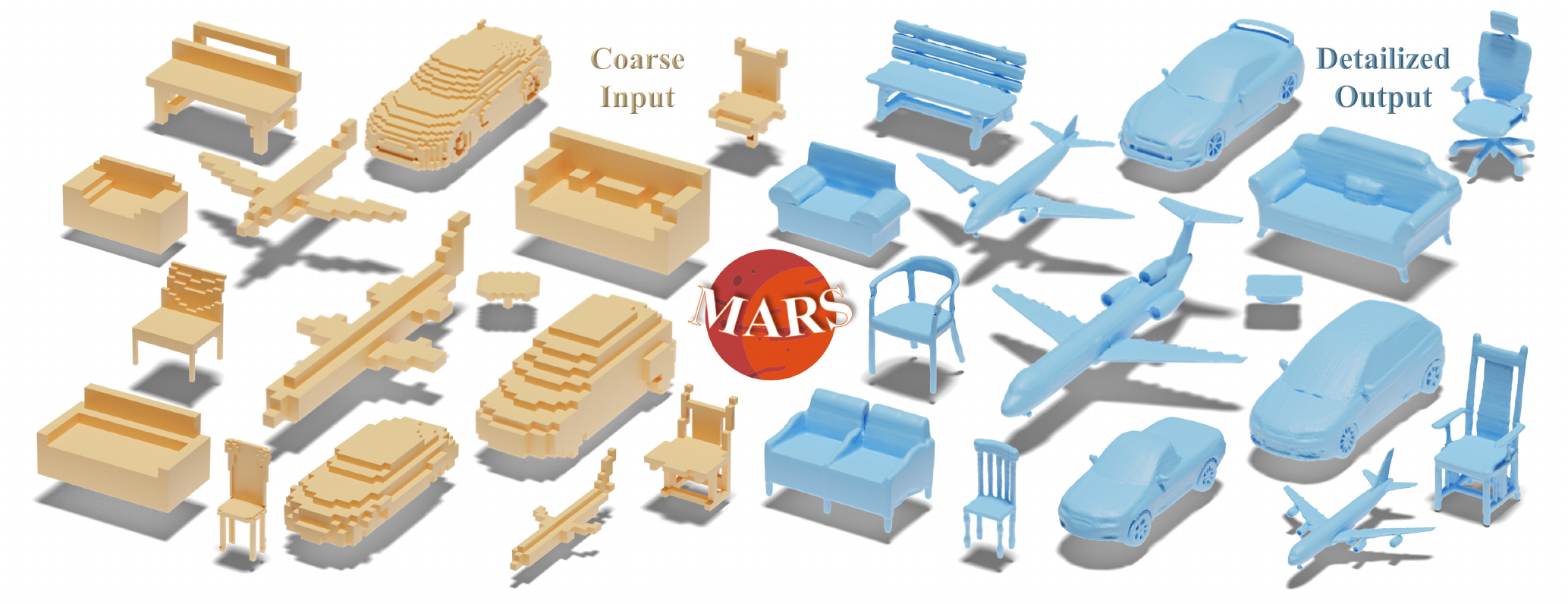}
        \caption{\textbf{A variety of objects generated by MARS.} The figure exemplifies the capability of MARS to generate intricate geometric patterns while preserving the overall shape integrity. }
\end{teaserfigure}

\maketitle

\section{Introduction}
With the rapid advancements in generative AI, 3D content has become increasingly prevalent and accessible. Recent developments in deep generative models, particularly those utilizing diffusion processes and vision-language architectures, have significantly enhanced both the accessibility and innovation of generated content. However, while these models excel at producing coarse content efficiently, they often fall short in generating high-quality geometric details. To overcome this limitation and create visually appealing 3D content with intricate geometry, some approaches have focused on scaling up models by training on larger datasets that encompass detailed geometric shapes. This strategy aims to push the boundaries of 3D content generation, ensuring that the resulting outputs are not only accessible but also rich in geometric detail and visual fidelity.

Nevertheless, directly generating shapes with the desired geometric details remains a formidable challenge, as it requires the generative model to grasp the intricate spatial structure of the shape, a task that is difficult to achieve without mesh supervision. Furthermore, in many real-world applications, obtaining a coarse shape is relatively straightforward, either by creating simple polygons or by utilizing similar templates from existing datasets. As a result, training a shape detailization model that transforms a coarse shape into a refined, detailed one emerges as a promising alternative for advancing 3D content generation.

Existing detailization methods predominantly employ Generative Adversarial Networks (GANs) to capture geometric details for shape refinement. These methods typically adhere to the Pyramid GAN structure, assigning different geometric details to various style codes. However, these style codes can only be effectively learned using datasets within similar categories (e.g., Chair and Desk). Consequently, constrained by the generalization capability of the style codes, previous detailization methods can only learn a limited range of styles and fail to fully utilize the available 3D data. This limitation hampers their ability to generalize across diverse categories and leverage the richness of existing 3D datasets, thereby restricting the scope and versatility of the generated geometric details.

To fulfill the task of 3D detailization while leveraging the wealth of available high-quality 3D data, we propose harnessing the power of 3D native generative models. Existing 3D native generative models predominantly rely on the diffusion transformer (DiT) structure for end-to-end generation. Consequently, applying coarse-to-fine techniques for detailization necessitates training multiple DiTs at different resolutions. This architecture demands substantial computational resources and meticulous structural design. In contrast, inspired by recent research in visual autoregressive models (VAR), our method employs an autoregressive model as the 3D shape detailization framework. Specifically, we encode mesh in different levels of detail (LODs) to different tokens, allowing a coarse shape to be interpreted as coarse-LOD tokens, while a fine shape corresponds to fine-LOD tokens. Through autoregressive generation, we achieve detailization by predicting next-LOD tokens given the coarse tokens as a condition.

Our mesh autoregressive model differs from 2D ones in two aspects: 1) 3D data necessitate a distinct encoding-decoding process compared to visual models. 2) The previous end-to-end supervision approach is inadequate for the task of detailization, as tokens of different scales fail to accurately perceive the level-of-detail information and maintain consistency across shapes at various scales. These aspects are crucial for the 3D generation task. Addressing these challenges is essential to ensure that the model can effectively capture and reproduce the intricate geometric details required for high-quality 3D shape detailization.
To address these issues, we introduce MARS, a Mesh AutoRegressive model for 3D Shape detailization. Our approach begins with a 3D shape encoding-decoding pipeline to train a multi-LOD, multi-category generative model. More specifically, we employ a 3D VQVAE with point cloud embedding to compress the mesh into compact tokens. During the training process, we design a geometry-consistency supervision technique to ensure that the tokens capture the geometry information across different levels of detail. This technique enables the model to learn the correspondence between multi-LOD tokens and meshes with different detailization levels. Following the multi-LOD 3D VQVAE, a sequence of discrete tokens is generated, which serves as the coarse-LOD input for the autoregressive model. MARS then accomplishes 3D shape detailization by predicting fine-LOD mesh tokens through next-LOD mesh token generation. Leveraging the capabilities of mesh autoregressive models, our method efficiently generates fine geometric details for various shape inputs while maintaining shape-consistency. Extensive experiments demonstrate the effectiveness of MARS in enhancing 3D shape detailization, showcasing its ability to produce intricate and high-quality geometric details.

To the best of our knowledge, MARS is the first method to leverage an autoregressive model for 3D shape detailization via next-LOD mesh token prediction. In summary, our proposed framework, MARS:
\begin{enumerate}
    \item Proposes a multi-LOD, multi-category 3D VQVAE network with 3D shape encoding-decoding to tokenize shapes into multi-LOD mesh tokens.
    \item Introduces a geometry-consistency supervision technique to enable the generative model to capture detailed geometric information across different levels of detail.
    \item Efficiently accomplishes high-quality 3D shape detailization through next-LOD mesh token prediction, achieving state-of-the-art performance across various benchmark settings.
\end{enumerate}

MARS represents a significant advancement in the field of 3D shape detailization, providing a robust and efficient solution for capturing and generating intricate geometric details.

\begin{figure*}
    \centering
    \includegraphics[width=0.975\linewidth]{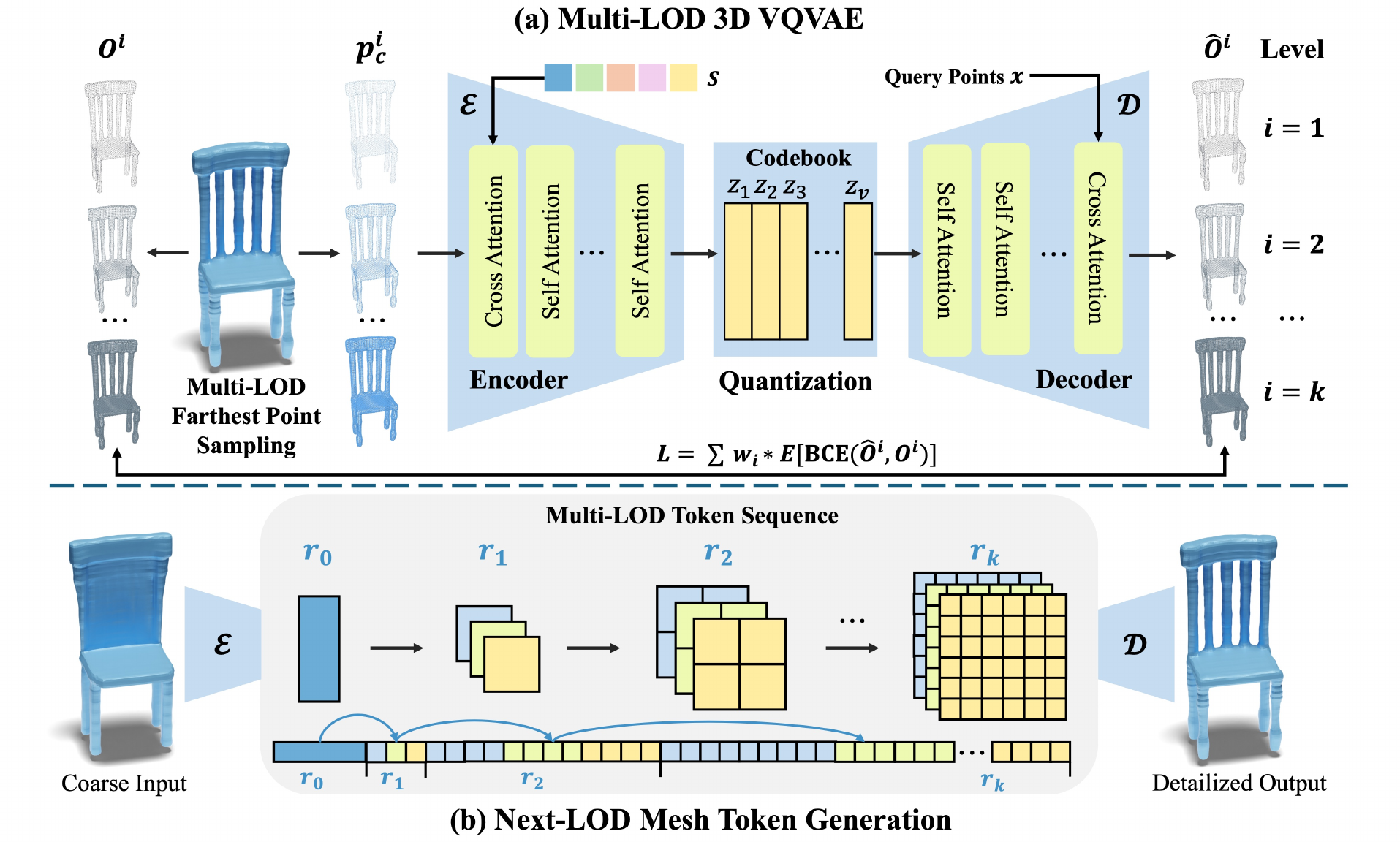}
    \caption{\textbf{Overview of MARS.} Our method initially employs a multi-LOD 3D Vector Quantized Variational Autoencoder (VQVAE) to tokenize input 3D meshes into discrete tokens representing multiple levels of detail. To effectively capture the geometric information across these varying levels, we have devised a geometry-consistency supervision strategy that enhances the training of the VQVAE. For the task of 3D shape detailization, we integrate a mesh autoregressive model that predicts next-LOD mesh tokens. Consequently, our model generates a detailized output that exhibits high-quality geometric details from a coarse input, thereby achieving sophisticated detail enhancement while maintaining structural integrity.}
    \label{fig:process}
\end{figure*}

\section{Related Work}
Our work is closely aligned with the domain of 3D generative models, specifically focusing on the intricate task of 3D shape detailization. Additionally, we delve into the realm of autoregressive models, as our method harnesses their capabilities to achieve high-fidelity mesh detailization.

\noindent\textbf{3D generative models.}
Capitalizing on the capabilities of variational autoencoders (VAEs)~\cite{vae,vqvae}, generative adversarial networks (GANs)~\cite{gan}, autoregressive models~\cite{pixelcnn,pixelcnnplus}, and diffusion probabilistic models~\cite{diffusion,DBLP:conf/icml/Sohl-DicksteinW15,DBLP:conf/iclr/0011SKKEP21}, a plethora of 3D generative models have been proposed for the generation of 3D shapes. These sophisticated deep generative models utilize voxels~\cite{DBLP:conf/nips/0001ZXFT16,DBLP:conf/eccv/ChoyXGCS16,xcube,DBLP:conf/3dim/HaneTM17}, point clouds~\cite{DBLP:conf/icml/AchlioptasDMG18,DBLP:conf/nips/zengVWGLFK22,craftsman,DBLP:conf/cvpr/FanSG17,DBLP:journals/tog/YinCHCZ19}, neural radiance fields~\cite{DreamFusion,magic3d,nerf,dreamgaussian,LGM,LRM}, or neural implicit representations~\cite{DBLP:conf/cvpr/GroueixFKRA18,pixel2mesh,DBLP:conf/iclr/ZhangCLGZ0F21,diffsdf,DBLP:conf/siggrapha/HuiLHF22,get3d,fantasia3d,neusdfusion,DBLP:conf/cvpr/ChenZ19,occnetwork,deepsdf,3dshape2vecset} to model 3D shapes. Among these methodologies, several have been developed to facilitate controllable 3D shape generation for modeling applications. For instance, Point-E~\cite{pointe}, Shap-E~\cite{shape}, and One-2-3-45~\cite{one2345} are capable of generating a 3D model from text or single-image inputs. Conversely, DECOR-GAN~\cite{decorgan}, ShaDDR~\cite{shaddr}, and DECOLLAGE~\cite{DECOLLAGE} have been designed to synthesize intricate 3D shapes from coarse voxel input. Despite the fact that DECOLLAGE~\cite{DECOLLAGE} facilitates interactive style control during generation, it is constrained by its ability to generate only a limited range of detailization styles and its inability to handle out-of-distribution inputs. In stark contrast, our proposed method harnesses the potential of large data and autoregressive models, thereby enabling the generation of a diverse array of detailization styles with superior details.

\noindent\textbf{3D shape detailization.}
Beyond the generation of 3D shapes from scratch, recent advancements have proposed methodologies for coarse-to-fine shape detailization, synthesizing geometric details in the process. Neural subdivision techniques~\cite{DBLP:journals/tog/LiuKCAJ20, 3DStyleNet} are designed to learn local geometric features from a reference 3D mesh, subsequently transferring these features to a novel shape. Mesh differentiable rendering methods~\cite{DBLP:journals/tog/LiuTJ18, Text2Mesh}, on the other hand, generate geometric details conditional on a reference image or text. However, these methods are limited by their inability to modify the coarse mesh topology, thereby constraining the range of synthesized detailization. To mitigate this limitation, mesh quilting~\cite{DBLP:journals/tog/ZhouHWTDGS06} adopts a strategy of copying and deforming local patches from a given geometric texture patch to detailize the mesh surface. DECOR-GAN~\cite{decorgan} utilizes the concept of image-to-image translation to generate detailed shapes from coarse voxels, given a conditioned geometric style. ShaDDR~\cite{shaddr} further enhances the quality of geometry by incorporating a 2-level hierarchical GAN. DECOLLAGE~\cite{DECOLLAGE} extends this by employing local style control for improved detailization quality. While these aforementioned methods support the generation of arbitrary mesh topology, they are confined to learned topology and thus, cannot generalize to diverse shapes. In contrast, MARS capitalizes on autoregressive models for next-LOD mesh token prediction, thereby enabling the detailization of a diverse array of shapes.

\noindent\textbf{Autoregressive models.}
Within the realm of image generation, early autoregressive models~\cite{pixelcnn,pixelcnnplus} were proposed to generate images as sequences of pixels. VQVAE~\cite{vqvae} and VQGAN~\cite{vqgan} introduced a strategy to quantize images into discrete tokens and employed a transformer to learn the autoregressive priors. Subsequent methods~\cite{maskgit,DBLP:journals/corr/abs-2406-07550} further enhanced the efficiency of tokenization. RQVAE~\cite{rqvae} incorporated multi-scale quantization to improve the quality of reconstruction, while VAR~\cite{var} proposed next-scale prediction for superior reconstruction and accelerated sampling speed. Concurrently, certain methods~\cite{DBLP:conf/icml/RameshPGGVRCS21,DBLP:journals/corr/abs-2406-06525} have endeavored to scale up autoregressive models in the task of text-conditioned image generation. However, the autoregressive approach has not been extensively explored in the field of 3D content generation. Our proposed method bridges this gap between autoregressive models and 3D generative models, achieving state-of-the-art performance in the benchmark for 3D shape detailization.

\section{Method}
We commence with a concise review of 3D native generative models and visual autoregressive models (VAR) to establish the foundational context requisite for this research (Section~\ref{sec:preliminaries}). To facilitate the detailization of 3D shapes, our framework, MARS, initially adopts a multi-LOD 3D vector-quantized variational autoencoder (VQVAE) (Section~\ref{sec:vqvae}) to encode input 3D models into multi-LOD discrete tokens. Subsequently, an autoregressive model is employed to generate mesh tokens at subsequent LOD settings (Section~\ref{sec:ar}). A comprehensive depiction of our methodology is illustrated in Fig.~\ref{fig:process}.

\subsection{Preliminaries}
\label{sec:preliminaries}
\noindent\textbf{3D native generative models}
With the advent of large-scale 3D datasets~\cite{objaverse,objaversexl}, recent research has focused on training 3D native generative models directly utilizing these datasets. These methodologies~\cite{shape,LN3Diff,craftsman,direct3d,3dshape2vecset,michelangelo,3dtopia,clay} predominantly employ a Variational Autoencoder (VAE)~\cite{vae} to compress 3D data into a compact latent representation and subsequently train latent diffusion models to synthesize such representations in latent space. In recent years, several approaches~\cite{meshanything,meshanything2,meshgpt} have been proposed to encode meshes into discrete tokens and learn to generate such token sequences using autoregressive transformers. Our method aligns with these 3D native generative models and employs a multi-LOD VQVAE to efficiently generate discrete tokens for autoregressive mesh generation.

\noindent\textbf{Visual autoregressive models (VAR)}
Previous visual autoregressive models~\cite{vqgan,DBLP:conf/cvpr/LeeKKCH22,DBLP:conf/iclr/YuLKZPQKXBW22} have predominantly relied on \textit{next-token} prediction to generate image sequences. To enhance sampling speed and reconstruction quality, VAR~\cite{var} introduces \textit{next-scale} token prediction for image generation.  Given an input image $I$, VAR initially encodes it into a feature map $f$ and quantizes $f$ into $K$ multi-scale token maps $R = \left(r_1, r_2, \ldots , r_K\right)$, where $r_K$ corresponds to the resolution of the input $f$. The autoregressive likelihood function is then formulated as:
\begin{equation}
    p\left(r_1, r_2, \ldots , r_K\right) = \prod_{k = 1}^K p\left(r_k|r_1, r_2, \ldots , r_{k-1}\right),
\end{equation}
where $r_k$ represents the token map at scale $k$. To facilitate next-scale prediction, VAR incorporates a VQVAE with a multi-scale quantizer $\mathcal{Q}(\cdot)$ to tokenize the input image into multi-scale discrete tokens $R$:
\begin{equation}
    f = \mathcal{E}(I), \mathrm{\quad} R = \mathcal{Q}(f),
\end{equation}
where $\mathcal{E}(\cdot)$ denotes the image encoder.
This quantization process map $f$ to a sequence of multi-scale token maps by identifying the nearest code in a codebook $Z$ containing $V$ vectors:
\begin{equation}
    z_k^{(i,j)} = (\mathrm{arg}\min_{v \in [V]} ||(\mathrm{lookup}(Z, v) - r_k^{(i,j)})|| ) \in [V],
\end{equation}
where $\mathrm{lookup}(Z, v)$ indicates the retrieval of the $v$-th vector from codebook $Z$. During the VQVAE training, each $z_k^{(i,j)}$ accesses $Z$ to approximate the original feature map $f$ as $\hat{f}$. A new image $\hat{I}$ is subsequently reconstructed using a decoder $\mathcal{D}(\cdot)$:
\begin{equation}
    \hat{f} = \mathrm{lookup}(Z, z), \mathrm{\quad} \hat{I} = \mathcal{D}(\hat{f}).
\end{equation}

\subsection{Multi-LOD 3D VQVAE with Geometry-Consistency Supervision}
\label{sec:vqvae}
As depicted in Fig.~\ref{fig:process}, our methodology commences by tokenizing a coarse mesh $m_c$ into a token map, followed by predicting a higher Level of Detail (LOD) token map $r_{d}$ using $r_c$ as prefix. The detailed mesh $m_{d}$ is then reconstructed by decoding  $r_{d}$. Analogous to visual autoregressive models\cite{var}, we employ a VQVAE for mesh tokenization. However, the conventional VQVAE architecture from VAR is not directly suitable for the task of detailization due to two primary reasons: \textbf{1) Distinct Encoding-Decoding Strategy for 3D Data:} Unlike VAR, which is optimized for 2D images, our approach is a 3D native generative model. This necessitates a specifically tailored encoder-decoder structure to accommodate the unique characteristics of 3D data. \textbf{2) Modified Supervision for Detailization:} VAR typically utilizes only the final-scale token for supervision. If applied directly using the original VQVAE structure, this would ensure the generation quality at the final scale but fail to maintain geometric consistency across intermediate scales, rendering it inadequate for comprehensive mesh detailization.

To overcome these challenges, we introduce a multi-LOD 3D VQVAE that incorporates geometry-consistency supervision. Initially, we utilize a 3D shape encoder-decoder framework to capture latent set representations. Subsequently, we integrate geometry-consistency supervision to effectively facilitate the task of detailization. This strategy ensures that each multi-LOD token contributes to the overall fidelity and geometric consistency of the generated meshes, aligning with the specific requirements of 3D data processing and detail enhancement. This approach not only enhances the detail fidelity but also ensures that the structural integrity of the mesh is maintained across different levels of detail.

\noindent\textbf{3D Shape Encoding-Decoding.}
Consistent with the methodologies outlined in~\cite{3dshape2vecset,craftsman,clay}, we transform 3D assets into a compact one-dimensional latent set, denoted as $\mathcal{S} = \{s_i \in \mathcal{R}^C\}_{i=1}^D$. In this notation, $D$ represents the number of latent sets, and $C$ indicates the dimensionality of the features within each set. We adopt an autoencoder architecture to encode 3D shapes into these latent sets. Specifically, for each 3D shape, we sample a set of point clouds, denoted as $p_c \in \mathcal{R}^{N \times 3}$, along with their corresponding surface normals $p_n$ where $N$  denotes the number of sampling points. We further apply Fourier positional encoding to capture the geometric information and integrate a cross-attention layer to infuse this information into the encoder. To effectively encapsulate geometric features, we utilize a Perceiver-based shape encoder $\mathcal{E}$, as proposed in~\cite{3dshape2vecset,craftsman}, to learn the latent set $S$:
\begin{equation}
\begin{aligned}
        p_f = \mathrm{PE}&(p_c) \oplus p_n, \\
    \mathcal{E}(p_c,p_n) = \mathcal{Q}(\mathrm{SelfAttn}&(\mathrm{CrossAttn}(S, p_f))), \\
\end{aligned}
\end{equation}
where $\oplus$ denotes concatenation, $\mathcal{Q}$ represents the quantizer, and $PE$ stands for the Fourier positional encoding function. To decode the latent set $S$, we employ a Perceiver-based decoder, similar to the one described in~\cite{craftsman}, which arranges all self-attention layers prior to the cross-attention layers. The decoder $\mathcal{D}$ predicts the occupancy value given a query 3D point $x$ and the shape latent embeddings $S$:
\begin{equation}
    O(\cdot) = O(\mathrm{CrossAttn}(\mathrm{PE}(x), S)),
\end{equation}
where $O(\cdot)$ is the occupancy prediction function using a Multilayer Perceptron (MLP). This architecture facilitates the accurate reconstruction of 3D shapes by leveraging the rich geometric information encoded in the latent sets.

\noindent\textbf{Geometry-Consistency Supervision.}
In the context of our mesh autoregressive models, we represent a detailed mesh as a finer Level of Detail (LOD) token, which is derived from a coarser-LOD token corresponding to the given coarse mesh. To achieve effective 3D shape detailization, our model is designed to capture the correspondence between multi-LOD tokens and meshes across varying levels of detail. Specifically, the prediction of the next-LOD token simulates the progressive mesh detailization process. We facilitate the learning of this pattern within the VQVAE by incorporating geometry-consistency supervision. For each point cloud $p$ in the training dataset, we initially generate a final-LOD shape through dense sampling and subsequently obtain a coarser-LOD shape via downsampling.
To manage geometric details across different LODs while maintaining overall geometric consistency, we utilize farthest point sampling (FPS) as our downsampling strategy. As the number of point clouds changes with each LOD after downsampling, the query 3D point is similarly processed, resulting in $x^i$ for the $i$-th LOD. This variation in point cloud density across LODs leads to latents of differing lengths, initially causing a dimension misalignment since the decoder requires a fixed-length latent input. To resolve this issue without the need for multiple LOD-specific decoders, we standardize the variable-length mesh tokens to a fixed-length sequence: 
\begin{equation}
    R^i = \mathrm{Pad}(R^i, 0),
\end{equation}
where we pad the original $i$-th LOD mesh token $R^i$ to the length of $R^K$. Following this procedure, we compute the occupancy $O^i$ for each LOD. 
This strategy simplifies the architecture by eliminating the need for multiple decoders and ensures that different LOD tokens effectively learn distinct levels of detail. The geometry-consistency supervision is then quantified using the Binary Cross Entropy (BCE) loss:
\begin{equation}
    \mathcal{L}_{vae} = \sum_{i=1}^K w_i \mathbb{E}_{x^i\in R^3}[\mathrm{BCE}(\hat{O}^i(x^i),O^i] + \lambda_{vq}\mathcal{L}_{vq},
\end{equation}
where $\hat{O}^i(x^i)$ represents the ground-truth occupancy value of the $i$-th LOD 3D point $x$, $w_i$ denotes the weight for the $i$-th LOD, $K$ is the final LOD number and $\mathcal{L}_{vq}$ is the commitment loss of the VQVAE following~\cite{vqvae} with weight $\lambda_{vq}$. This formulation ensures that our model robustly learns detailed geometric features across multiple LODs, enhancing the fidelity and consistency of the generated 3D meshes.

\begin{figure*}
    \centering
    \includegraphics[width=0.95\linewidth]{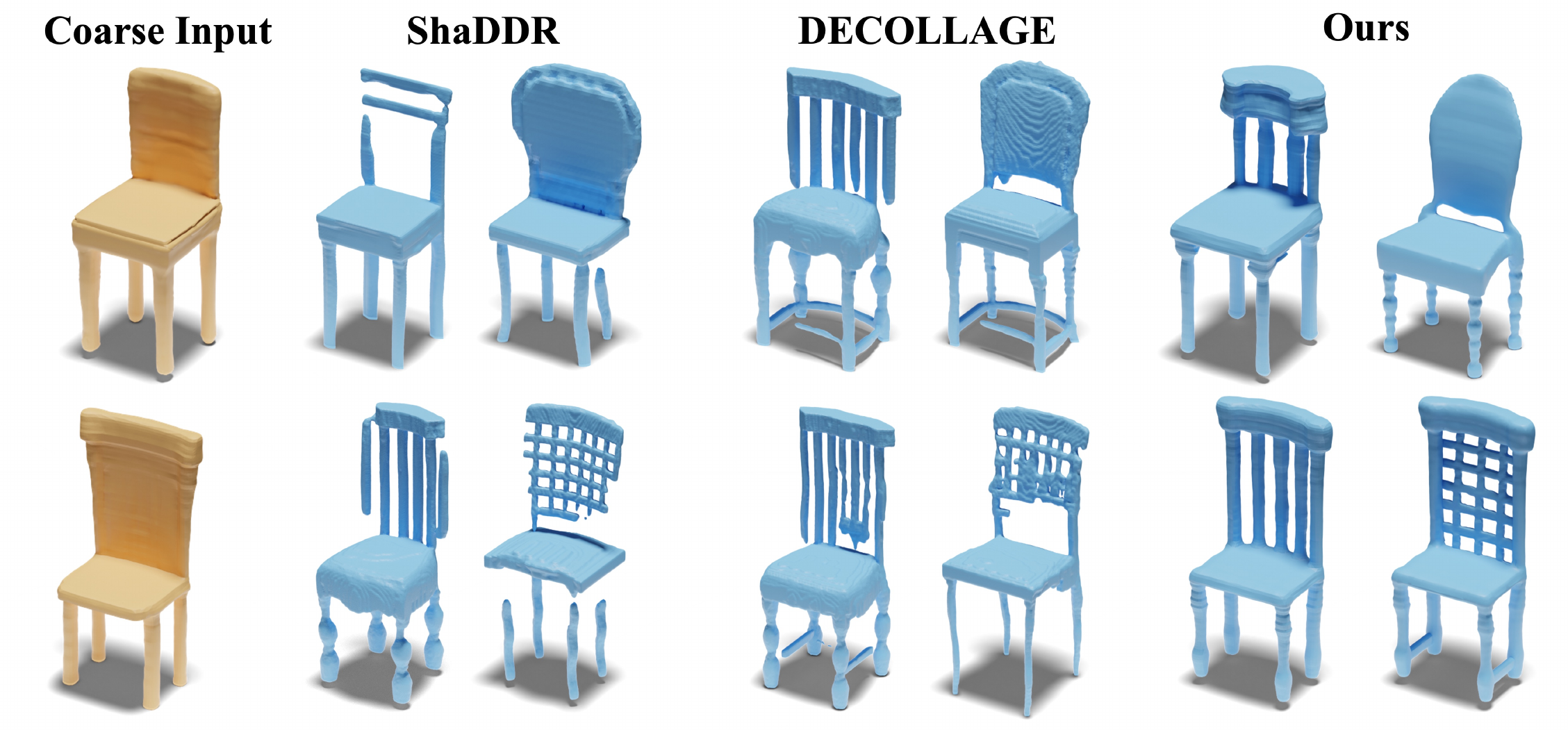}
    \caption{\textbf{Comparison with previous detailization approaches.} We conduct a comparative analysis of our model with existing detailization methods, specifically ShaDDR and DECOLLAGE. For each coarse input, we demonstrate two distinct detailization styles. It is observed that both ShaDDR and DECOLLAGE often produce outputs with compromised mesh integrity. In contrast, our method consistently generates complete outputs that exhibit high-quality geometric details, thereby underscoring the robustness and efficacy of our approach in handling complex detailization tasks.}
    \label{fig:comp}
\end{figure*}

\subsection{Next-LOD Mesh Token Generation}
\label{sec:ar}
Upon training the multi-LOD 3D VQVAE, a sequence of discrete tokens is generated, which then serves as the input for the autoregressive model. To facilitate the prediction of the next-LOD mesh token for 3D shape detailization, we employ an autoregressive (AR) model that utilizes a decoder-only transformer architecture, similar to those used in GPT-2 and VQGAN, and incorporates adaptive normalization (AdaLN). The autoregressive process initiates with a single token and incrementally predicts subsequent, finer LOD tokens, each conditioned on the preceding coarser tokens. To enhance computational efficiency, all tokens $L^{(s)}$ at LOD $s$ are generated simultaneously in parallel. During the training phase, a block-wise causal attention mask is employed to ensure that each token only attends to its preceding tokens, thereby preserving the causal structure necessary for the autoregressive model. For the inference phase, we integrate the kv-caching technique~\cite{DBLP:conf/mlsys/PopeDCDBHXAD23} to expedite the sampling process. This approach not only streamlines the generation of high-fidelity 3D meshes but also ensures that the detailization process is both efficient and scalable.

\section{Experiment}

\subsection{Mesh Detailization Comparison}
\begin{figure}
    \centering
    \includegraphics[width=0.95\linewidth]{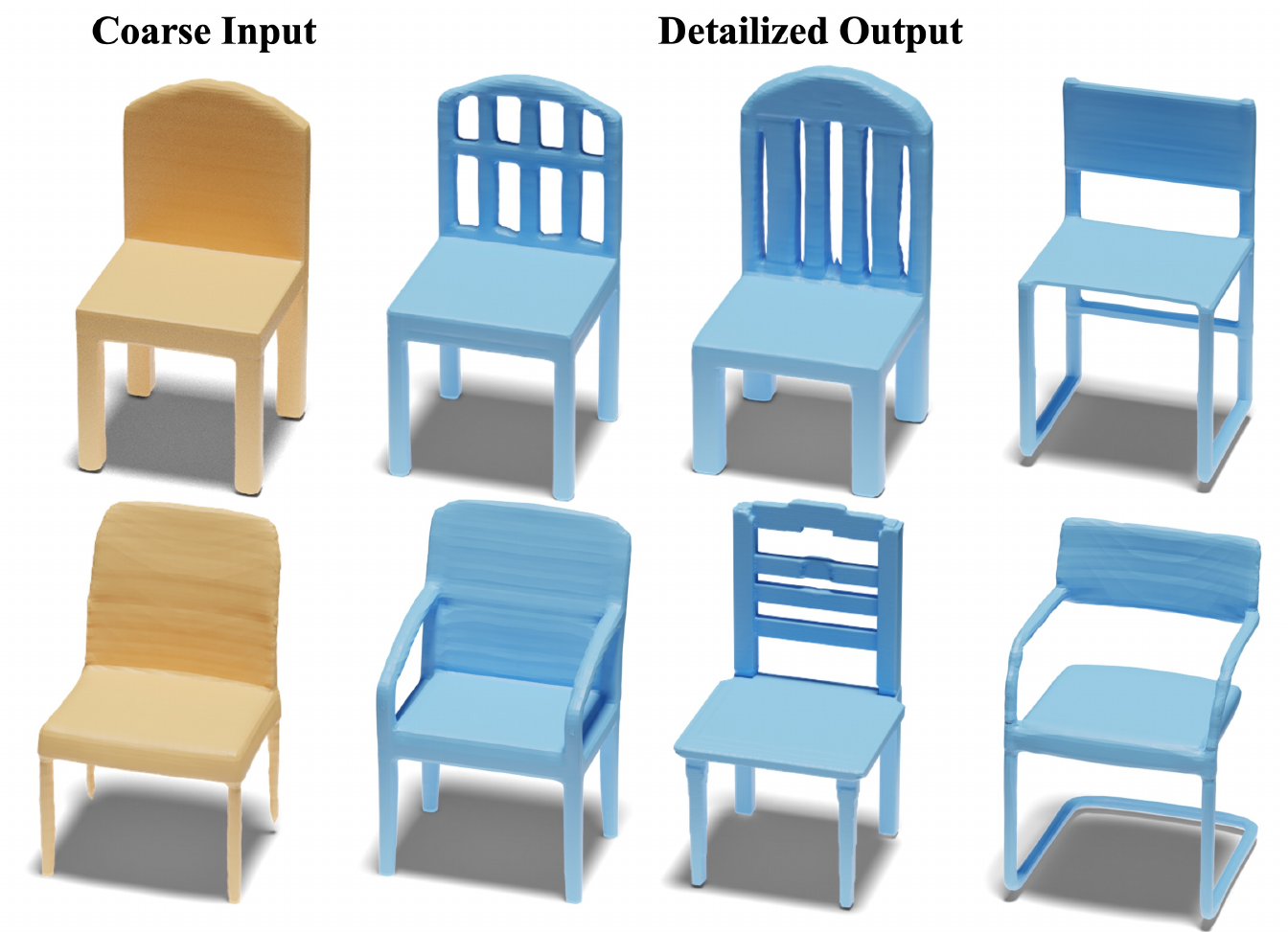}
    \caption{\textbf{Diverse generation results.} Our approach is capable of generating a diverse array of detailed meshes from a uniform coarse input. This capability underscores the robustness and adaptability of our method in enhancing mesh detailization across various scenarios.}
    \label{fig:d}
\end{figure}

\begin{figure*}
    \centering
    \includegraphics[width=0.9\linewidth]{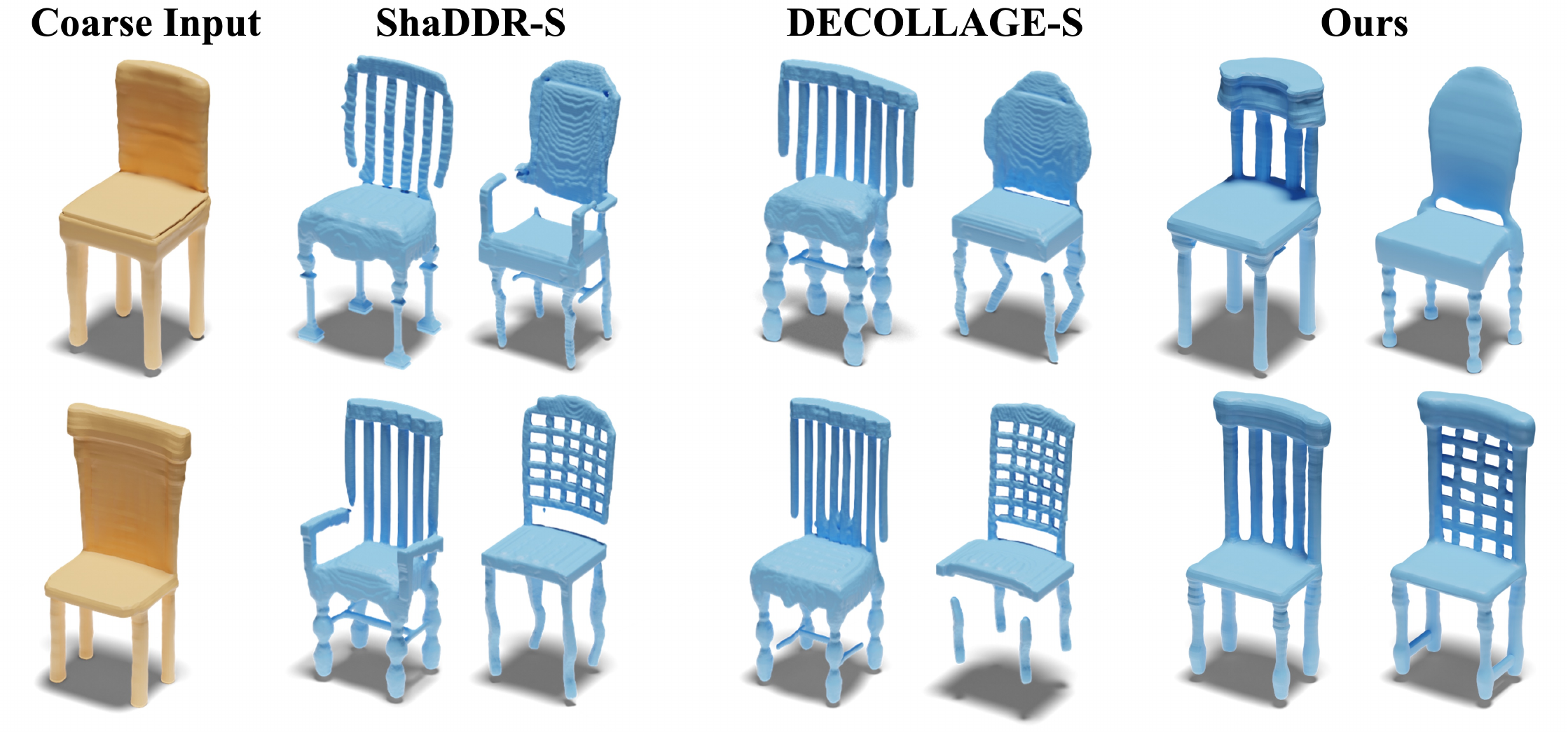}
    \caption{\textbf{Comparison with previous detailization methods.} We evaluate our model by comparing it with the original ShaDDR and DECOLLAGE, both trained using datasets from similar categories. For each coarse input, we present two distinct detailization styles, demonstrating the versatility and adaptability of our approach in generating refined outputs.}
    \label{fig:single}
\end{figure*}

\begin{figure*}
    \centering
    \includegraphics[width=0.855\linewidth]{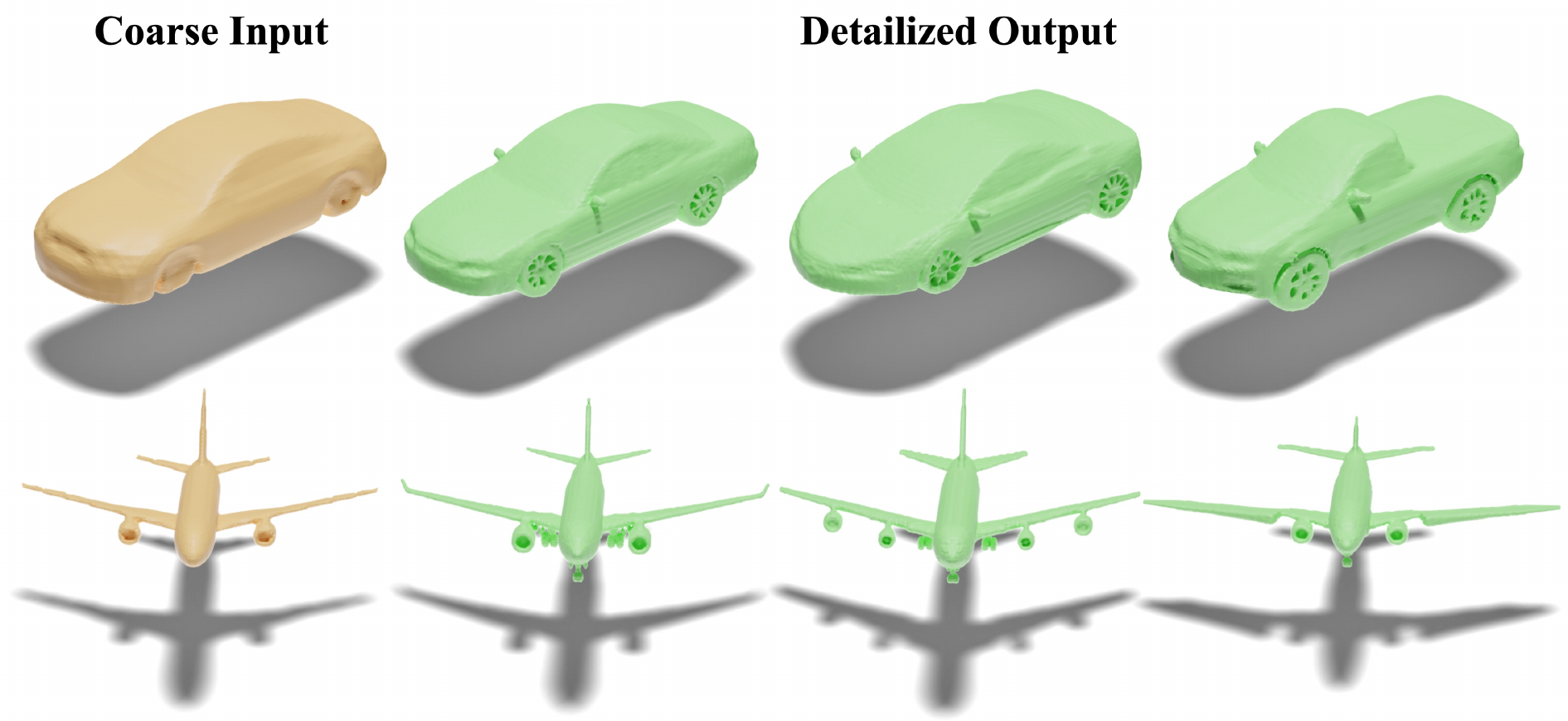}
    \caption{\textbf{Diverse generation results.} Our method is capable of generating a diverse array of detailed results from the same coarse input, showcasing its robustness and flexibility in handling variations in 3D shape refinement.}
    \label{fig:d2}
\end{figure*}

\noindent\textbf{Qualitative Comparison.}
We evaluate our model against state-of-the-art mesh detailization approaches, namely ShaDDR~\cite{shaddr} and DECOLLAGE~\cite{DECOLLAGE}, using the benchmark dataset previously described. For a consistent evaluation framework, we adopt the methodology from prior works\cite{shaddr,DECOLLAGE}, utilizing a template mesh for each category as the input and applying different detailization methods to ensure comparability. Fig.~\ref{fig:comp} presents qualitative comparisons with these two leading detailization approaches. As illustrated, both ShaDDR and DECOLLAGE exhibit limitations in generating detailed geometry when trained on multi-category data. These methods, which employ GANs for detailization, necessitate training datasets comprised of similar categories due to their reliance on learning category-specific style codes for detailization. Consequently, their inability to generalize across a multi-category dataset restricts their flexibility and applicability. Additionally, these methods often produce damaged mesh parts, further impeding practical applications. In contrast, our approach harnesses the capabilities of the geometry-consistent VQVAE model, adept at representing multi-category meshes across various levels of detail. The 3D autoregressive model we propose effectively generates multi-LOD representations across different categories and detail levels, thereby demonstrating superior performance compared to the baseline models. Furthermore, we juxtapose our detailization results with those of ShaDDR and DECOLLAGE trained on similar-category data. As depicted in Fig.~\ref{fig:single}, our method not only preserves complete shape integrity but also enhances the geometric detail, surpassing previous methods. Additionally, Figs.~\ref{fig:d} and~\ref{fig:d2} showcase the diversity of our generation results. These figures illustrate that our method is capable of producing a variety of detailed outputs from the same coarse input, highlighting the robustness and versatility of our approach in generating diverse and detailed 3D meshes.

\noindent\textbf{Quantitative Comparison.}
To rigorously evaluate the effectiveness of our model, we adopt several quantitative metrics as utilized in prior research. We conduct comparisons of our approach with ShaDDR and DECOLLAGE across both multi-category datasets, which highlight the limitations of baseline models in handling objects with significant variations, and similar-category datasets, denoted as \textit{ShaDDR-S} and \textit{DECOLLAGE-S} in accordance with the literature, to demonstrate the consistent superiority of our model over the baseline approaches. All models are evaluated following the established protocol of DECOLLAGE, and the results are presented in Table~\ref{tab:diverse}. 
To determine the fidelity with which the generated shapes adhere to the structures of the input shapes, we employ the Strict-IOU metric, which quantifies the Intersection over Union (IOU) between the downsampled detailized output and the coarse input. Additionally, we utilize the Loose-IOU metric to calculate the proportion of occupied voxels in the input that correspond to similarly occupied voxels in the downsampled output. 
Furthermore, we employ the F-Score to assess the similarity between the generated results and the coarse input shapes at a local level. A high F-Score indicates that the local details of the generated shapes closely mirror those of the input shapes, thereby enhancing the local plausibility of the generated shapes.

\begin{table}
    \centering
    \begin{tabular}{c|ccc}
        \hline
         Method & Strict-IOU $\uparrow$ & Loose-IOU $\uparrow$ & F-Score $\uparrow$ \\
         \hline
         ShaDDR &  0.512  &    0.668     &  0.898    \\
         \textit{ShaDDR-S} &  0.596  &    0.760     &  0.907    \\
         DECOLLAGE &  0.627    &   0.791      &  0.901    \\
         \textit{DECOLLAGE-S} &  0.748    &   0.908      &  0.914     \\
         Ours &  \textbf{0.785}   &    \textbf{0.924}    &  \textbf{0.918}   \\
         \hline
    \end{tabular}
    \caption{\textbf{Quantitative comparison of different detailization methods.} Our approach consistently surpasses previous methodologies across all evaluation metrics, within both multi-category and similar-category settings. For clarity, ShaDDR and DECOLLAGE when applied to similar-category datasets are referred to as \textit{ShaDDR-S} and \textit{DECOLLAGE-S}, respectively.}
    \label{tab:diverse}
\end{table}

\subsection{Ablation Study}
\begin{figure}
    \centering
    \includegraphics[width=\linewidth]{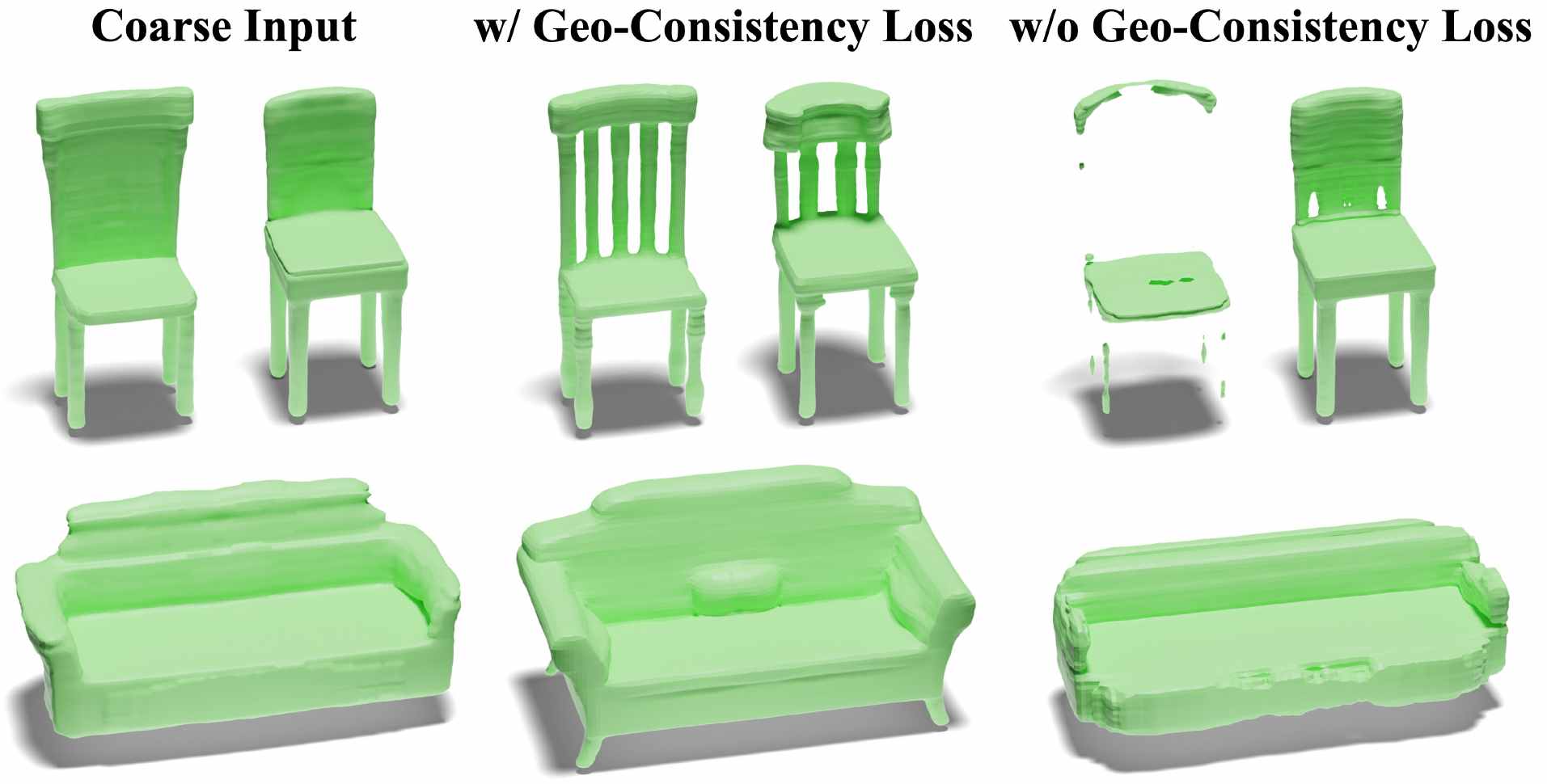}
    \caption{\textbf{Reconstruction ablation of geometry-consistency supervision.} It can be seen that the geometry-consistency supervision is crucial in our shape detailization process.}
    \label{fig:ms_sup}
\end{figure}

\noindent\textbf{Geometry-Consistency Supervision.}
Our method integrates geometry-consistency supervision to ensure shape-consistent mesh generation across varying levels of detail. The efficacy of this supervision is demonstrated through the presentation of ablation reconstruction results in Fig.~\ref{fig:ms_sup}. Our model facilitates detailization by initially representing the coarse mesh input as a coarse token, followed by the prediction of subsequent level-of-detail (LOD) mesh tokens. The geometry-consistency supervision acts as a directive for LOD guidance within the mesh token framework. As evidenced, the incorporation of geometry-consistency supervision enables our model to incrementally learn the detailization process. In contrast, the absence of geometry-consistency supervision results in the reconstruction of compromised geometries, as the model struggles to interpret the LOD nuances associated with different mesh tokens.

\begin{figure}
    \centering
    \includegraphics[width=\linewidth]{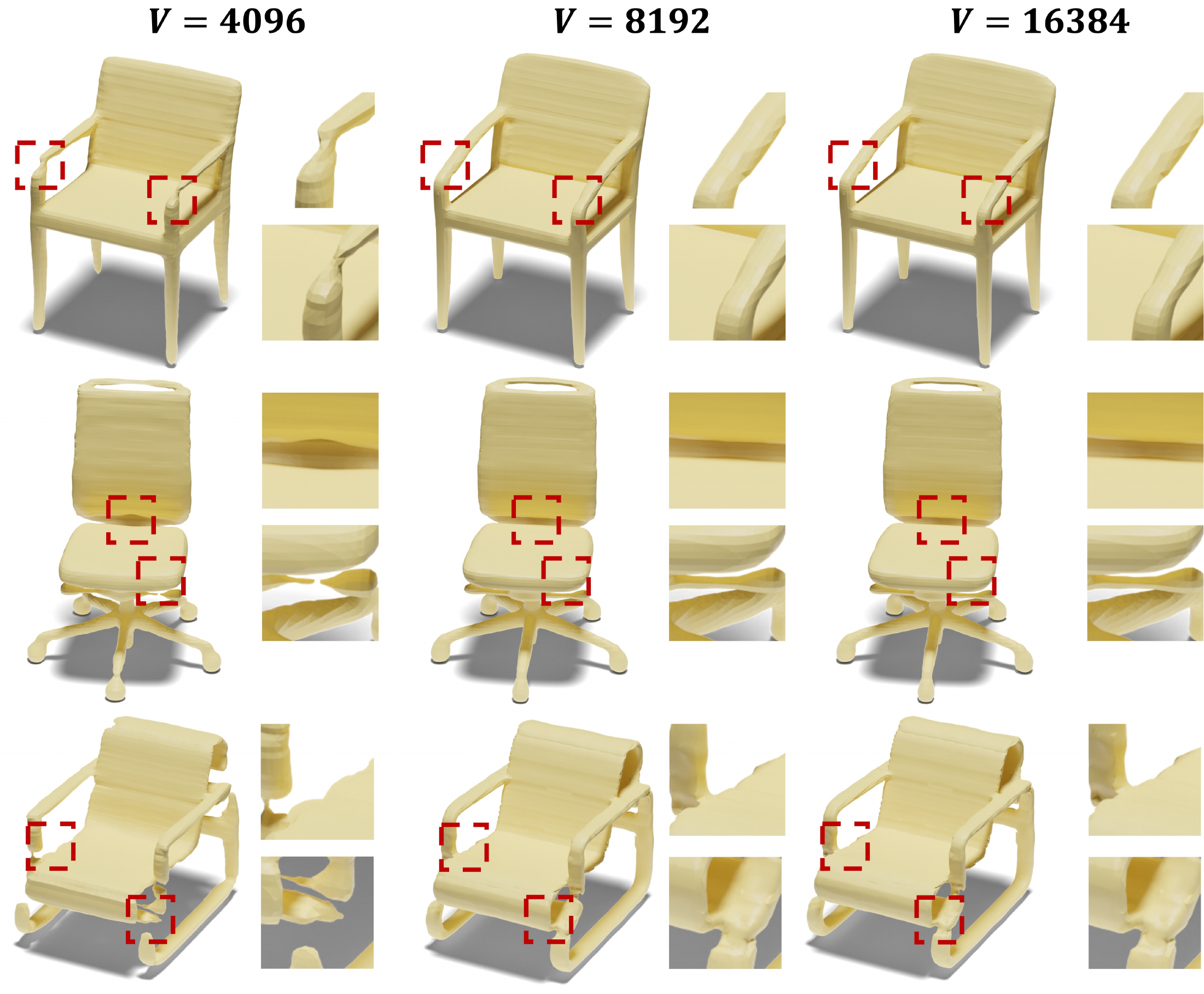}
    \caption{\textbf{Comparison of reconstruction using different codebook sizes.} It is apparent that our method yields comparable outcomes when employing codebook sizes of 8192 and 16384, with these results surpassing those obtained using a codebook size of 4096.}
    \label{fig:cb}
\end{figure}

\noindent\textbf{VQVAE Codebook Size.}
During the training of the VQVAE, the size of the codebook significantly influences the quality of the reconstruction results. We conduct ablation studies by varying the codebook size $V$ to 4096, 8192, and 16384. The comparative results are illustrated in Fig.~\ref{fig:cb}. To quantitatively evaluate the reconstruction quality, we measure the accuracy of the recovered occupancy and the corresponding Intersection Over Union (IOU) derived from the latents on a subset of our constructed dataset. These results are detailed in Table~\ref{tab:codebook}. Our method achieves comparable performance when utilizing codebook sizes of 8192 and 16384, which are superior to the results obtained with a codebook size of 4096.

\begin{table}
    \centering
    \begin{tabular}{c|ccc}
        \hline
         Codebook Size $V$ & 4096 & 8192 &16384 \\
         \hline
         Accuracy &  0.902  &    0.964     &  0.972   \\
         IOU & 0.874 & 0.904  & 0.916 \\ 
         \hline
    \end{tabular}
    \caption{\textbf{Quantitative comparison of different codebook sizes $V$.} The results demonstrate that utilizing a codebook size of 8192 yields superior outcomes compared to a size of 4096, and achieves results that are comparable to those obtained with a codebook size of 16384.}
    \label{tab:codebook}
\end{table}

\begin{figure*}
    \centering
    \includegraphics[width=0.9\linewidth]{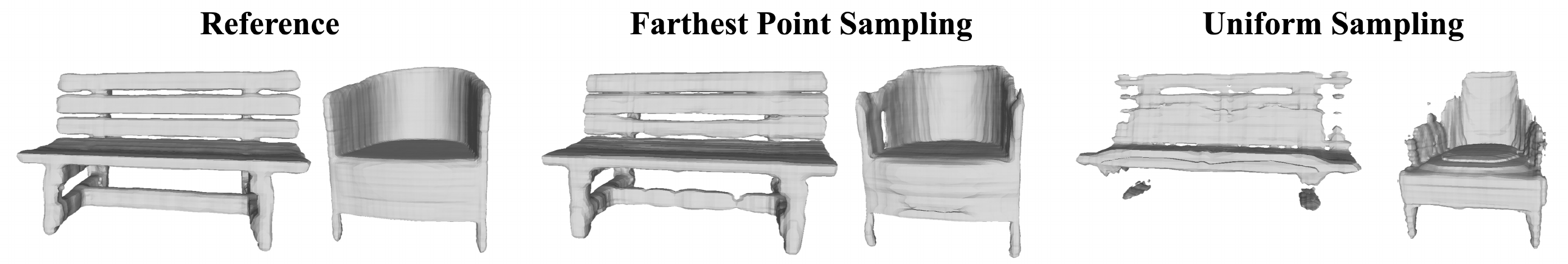}
    \caption{\textbf{Downsampling strategy ablation.} We present the mid-Level of Detail (LOD) reconstruction results utilizing both sampling strategies, with the final-LOD reconstruction serving as a reference for comparative analysis.}
    \label{ds}
\end{figure*}

\noindent\textbf{Downsampling Strategy.}
By incorporating a geometry-consistency supervision loss, our method effectively captures geometric information across various levels of detail. During training, we downsample the point clouds to create a dataset that encompasses different levels of geometric detail. The farthest point sampling (FPS) strategy is utilized for downsampling. We conduct an ablation study to compare the efficacy of the FPS strategy with that of uniform sampling. The results, including mid-Level of Detail (LOD) reconstructions using each sampling strategy and the final-LOD reconstruction as a reference, are presented. As illustrated in Fig.~\ref{ds}, our model, when employing FPS, more effectively captures the level-of-detail information. This effectiveness is attributed to the FPS's tendency to preserve the overall shape of the geometry during downsampling. Consequently, each downsampling step primarily reduces high-frequency geometric details while retaining the structural integrity of the shape. The model subsequently learns to incrementally refine the geometric details in a reverse manner.

\subsection{Applications}
\begin{figure*}
    \centering
    \includegraphics[width=\linewidth]{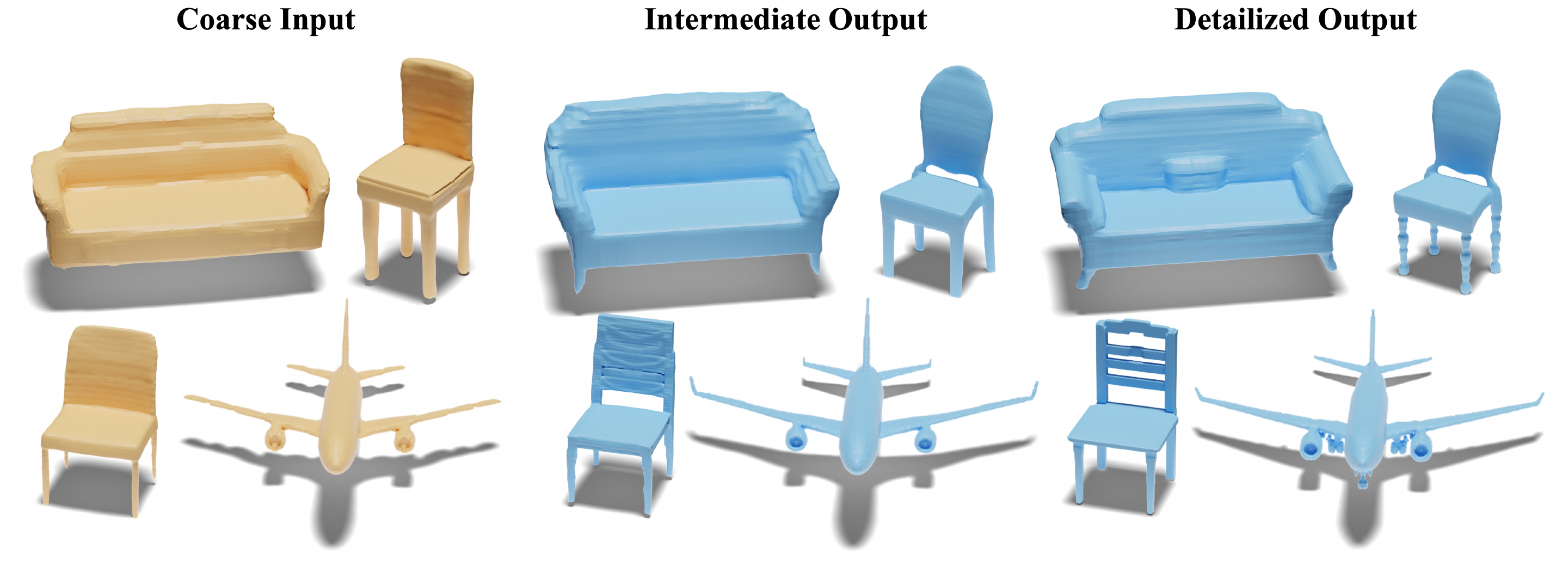}
    \caption{\textbf{Multi-LOD detailized output.} We demonstrate the detailization capabilities of our model by showcasing the generation of meshes across various scales. This process highlights the model's proficiency in producing detailed geometric structures from coarse to fine LODs.}
    \label{fig:process_vis}
\end{figure*}

\begin{figure*}
    \centering
    \includegraphics[width=0.95\linewidth]{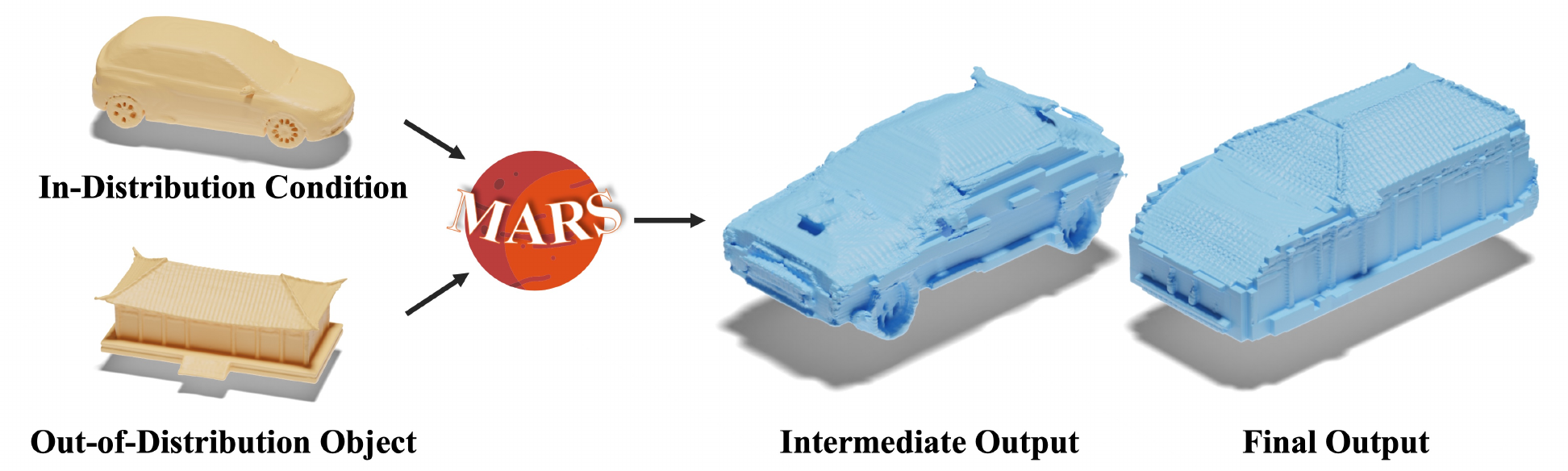}
    \caption{\textbf{Mixing styles application.} Our model is capable of performing style-mixing by utilizing one shape as a conditioning input to detailize another shape. It underscores the model's flexibility in integrating stylistic elements from one geometry into the detailed enhancement of another.}
    \label{fig:app}
\end{figure*}

\begin{figure*}
    \centering
    \includegraphics[width=\linewidth]{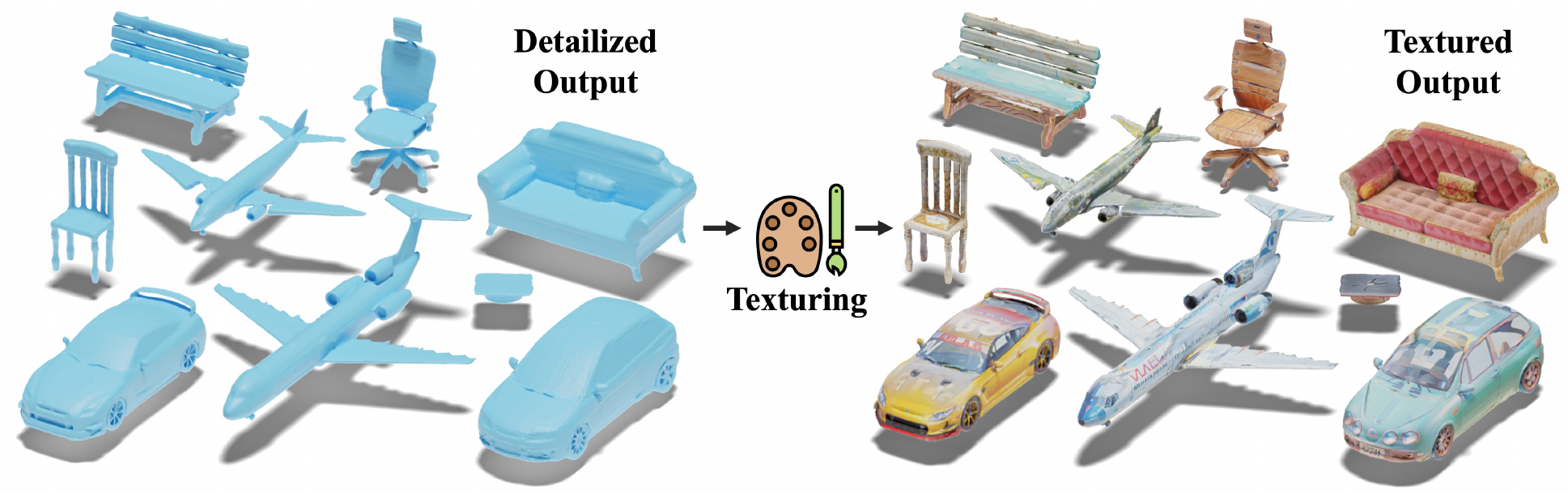}
    \caption{\textbf{Textured results using detailized output.} We demonstrate the results of texturing the detailized output in the teaser section, showcasing the enhanced visual fidelity achieved through our proposed methodology.}
    \label{fig:textured}
\end{figure*}

Our model enhances a coarse shape by tokenizing it into mesh tokens and predicting the next Level of Detail (LOD) mesh tokens. Consequently, our approach is capable of constructing meshes with varying levels of detail, as depicted in Fig.\ref{fig:process_vis}. We further evaluate our model on an out-of-distribution object (a house from ShaDDR~\cite{shaddr}) and provide a comparison with an in-distribution shape condition (a car from ShapeNet~\cite{shapenet}). The results, illustrated in Fig.\ref{fig:app}, demonstrate that our method effectively generates a house in the form of a car. This application underscores our model's ability to capture geometric features and generate high-quality details while maintaining the overall shape integrity. Additionally, we present the results of texturing the refined shape in Fig.~\ref{fig:textured}. Given that our model processes a coarse mesh and outputs a detailed mesh, it can serve as a valuable tool for mesh refinement within the 3D generation pipeline, offering an efficient and robust method for capturing and rendering intricate geometric details.

\section{Conclusion}
In this work, we introduce MARS, a mesh autoregressive model designed for 3D shape refinement. Our method leverages the capabilities of autoregressive models to generate high-fidelity geometric details. MARS is founded on two principal components: 1) A multi-Level of Detail (LOD) 3D Vector Quantized Variational Autoencoder (VQVAE) that encodes and decodes 3D shapes into multi-LOD mesh tokens, and 2) An autoregressive mechanism for predicting next-LOD mesh tokens, facilitating high-quality 3D shape refinement. Extensive experiments validate that our approach not only achieves high-quality refinement but also preserves the overall shape integrity. Our model sets a new benchmark in the domain of 3D shape refinement, demonstrating state-of-the-art performance and marking a significant advancement in the field.

\newpage

\bibliographystyle{ACM-Reference-Format}
\bibliography{sample-bibliography}

\end{document}